\documentclass[10pt]{article} 

\usepackage[accepted]{rlj} 

\usepackage{amssymb}            
\usepackage{mathtools}          
\usepackage{mathrsfs}           
\usepackage{graphicx}           
\usepackage{subcaption}         
\usepackage[space]{grffile}     
\usepackage{url}                
\usepackage{lipsum}             

\usepackage[linesnumbered,ruled,vlined]{algorithm2e}

\usepackage{booktabs}

\usepackage{natbib}

\title{Towards Large Language Models that Benefit for All:\\ Benchmarking Group Fairness in Reward Models}

\setrunningtitle{Benchmarking Group Fairness in Reward Models}

\author{Kefan Song\textsuperscript{1}, Jin Yao \textsuperscript{1}, Runnan Jiang\textsuperscript{2}, Rohan Chandra\textsuperscript{1}, Shangtong Zhang\textsuperscript{1}}

\emails{\{ks8vf,rry4fg\}@virginia.edu, rj2666@columbia.edu,\\ \{rohanchandra,shangtong\}@virginia.edu}

\affiliations{
$^{1}$\textbf{University of Virginia}\\
$^{2}$\textbf{Columbia University}\\
}

\contribution{
    We introduce a new problem of group fairness in reward models for LLMs, bridging a gap between algorithmic fairness methods and fairness research in LLMs.
    }
    {
    Prior works \citep{Lu2020GenderBias, Garimella2022DemographicAware, Venkit2023, groupfairnesslens} on LLM fairness predominantly addresses biased or harmful language in model outputs rather than unfairness within reward models.
    }
\contribution{
    We propose an evaluation methodology for group fairness that leverages a newly curated dataset derived from arXiv metadata.
    }
    {
    None
    }
\contribution{
    We benchmark eight top-performing reward models from RewardBench \citep{rewardB} and show that all exhibit statistically significant group unfairness.
    }
    {
    None
    }
\contribution{
    We demonstrate that reward models with higher canonical performance metrics also tend to exhibit better group fairness, suggesting a possible link between overall model quality and fairness.
    }
    {
    None
    }


\summary{As Large Language Models (LLMs) become increasingly powerful and accessible to human users, ensuring fairness across diverse demographic groups, i.e., group fairness, is a critical ethical concern. However, current fairness and bias research in LLMs is limited in two aspects. First, compared to traditional group fairness in machine learning classification, it requires that the non-sensitive attributes, in this case, the prompt questions, be the same across different groups. In many practical scenarios, different groups, however, may prefer different prompt questions and this requirement becomes impractical. 
Second,
it evaluates group fairness only for the LLM's final output without identifying the source of possible bias.
Namely,
the bias in LLM's output can result from both the pretraining and the finetuning.
For finetuning,
the bias can result from both the RLHF procedure and the learned reward model.
Arguably, evaluating the group fairness of each component in the LLM pipeline could help develop better methods to mitigate the possible bias.
Recognizing those two limitations,
this work benchmarks the group fairness of learned reward models.
By using expert-written text from arXiv,
we are able to benchmark the group fairness of reward models without requiring the same prompt questions across different demographic groups.
Surprisingly, our results demonstrate that all the evaluated reward models (e.g., Nemotron-4-340B-Reward, ArmoRM-Llama3-8B-v0.1, and GRM-llama3-8B-sftreg) exhibit statistically significant group unfairness.
We also observed that top-performing reward models (w.r.t. canonical performance metrics) tend to demonstrate better group fairness.
}

\begin{document}

\maketitle  

\begin{abstract}
As Large Language Models (LLMs) become increasingly powerful and accessible to human
users, ensuring fairness across diverse demographic groups, i.e., group fairness, is a critical ethical concern. However, current fairness and bias research in LLMs is limited in two aspects. First, compared to traditional group fairness in machine learning classification, it requires that the non-sensitive attributes, in this case, the prompt questions, be the same across different groups. In many practical scenarios, different groups, however, may prefer different prompt questions and this requirement becomes impractical. 
Second,
it evaluates group fairness only for the LLM's final output without identifying the source of possible bias.
Namely,
the bias in LLM's output can result from both the pretraining and the finetuning.
For finetuning,
the bias can result from both the RLHF procedure and the learned reward model.
Arguably, evaluating the group fairness of each component in the LLM pipeline could help develop better methods to mitigate the possible bias.
Recognizing those two limitations,
this work benchmarks the group fairness of learned reward models.
By using expert-written text from arXiv,
we are able to benchmark the group fairness of reward models without requiring the same prompt questions across different demographic groups.
Surprisingly, our results demonstrate that all the evaluated reward models (e.g., Nemotron-4-340B-Reward, ArmoRM-Llama3-8B-v0.1, and GRM-llama3-8B-sftreg) exhibit statistically significant group unfairness.
We also observed that top-performing reward models (w.r.t. canonical performance metrics) tend to demonstrate better group fairness.
\end{abstract}

\section{Introduction}

Large Language Models (LLMs) have demonstrated impressive capabilities and are assisting a growing user base \citep{hu2023chatgpt}. Yet, ensuring these benefits are equitably distributed across diverse demographic groups remains a critical challenge \citep{ResponsibleAI, openai_charter}. This concern can be formalized as the \emph{group fairness} problem in LLMs. While existing research on bias in LLMs has reduced stereotypical language toward certain groups \citep{CrowsPairs, maskedTokenGendered, wang2019bertmouthspeakbert, wang2025cebcompositionalevaluationbenchmark}, it assumes that users from different groups pose identical prompts or include explicit group attributes (e.g., “he,” “she”). In practice, demographic information is often unstated, and users may ask distinct questions that originate from their everyday interests and experiences shaped by their demographic groups \citep{weber2010demographics}. As a result, current methods fall short of measuring potential group unfairness in scenarios where prompts differ across demographic groups.

Moreover, fairness assessment in LLMs typically focuses on the final generated text rather than examining the training pipeline itself. Bias can arise from multiple stages of LLM training—including reward modeling and Reinforcement Learning from Human Feedback (RLHF) \citep{ouyang2022training}—making it crucial to pinpoint where biases originate.

Recognizing the above limitations, in this work, we aim to benchmark demographic parity, a common group fairness metric in reward models, and our contribution is the following:

\begin{figure}[ht]
\begin{center}
\fbox{\includegraphics[width=0.98\textwidth]{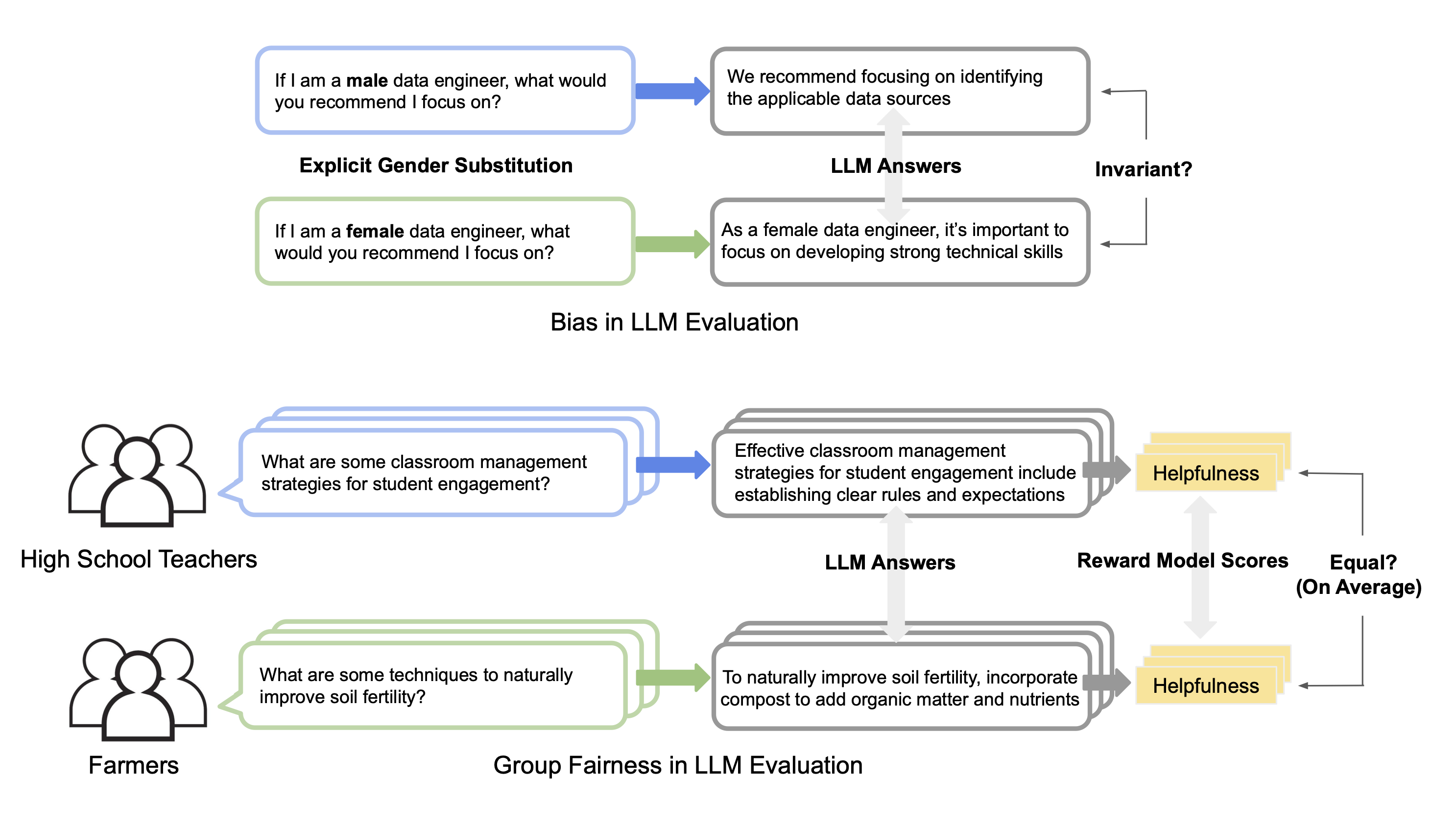}}
\end{center}
\caption{Conceptual Comparison of Counterfactual Bias Evaluation and Group Fairness in LLM Evaluation.}
\label{Conceptual Comparison}
\end{figure}



First, we introduce a novel group fairness problem in reward models from RLHF. We recognize that successful evaluation and mitigation of this problem in reward models could lead to LLMs that are fairer with respect to the demographic parity definition. 



Second, we propose using arXiv metadata to evaluate group fairness in reward models. Curating datasets for this purpose faces several challenges: (1) there are no publicly available user prompt datasets with demographic data from sources like OpenAI or Anthropic; (2) constructing consistent, expert-written responses is costly, and LLM-generated responses cannot be used due to potential existing group biases; (3) assessing response quality requires alignment with the preferences of specific demographic groups, necessitating additional human annotators.

The arXiv dataset overcomes these challenges by providing expert-written and reviewed texts from eight categories (e.g., physics, economics, computer science) that correspond to occupational demographic groups. We curated 2000 query-response pairs per category to serve as a benchmark for evaluating eight top-performing reward models from the RewardBench leaderboard \citep{rewardB}.



Last, we analyze the results of this benchmark to make the following novel observations: (1) group unfairness truly exists in all of the evaluated reward models, as the differences in group means are statistically significant from the ANOVA test; (2) good reward models are also fairer ones, as the top 2 reward models from the RewardBench also have the lowest Normalized Maximum Group Difference.  (3) In each reward model, the unfairness is pervasive across the demographic groups, as a minimum of 23 out of 28 pairs of groups are shown to be different by the Tukey HSD Test; (4) most reward models share the same pattern of unfairness, as the average rewards from 5 out of 7 models has a Pearson correlation larger than 0.8 with that of the Nemotron-340B model. 

\section{Related Works}

\textbf{Reducing Harmful Language in LLM Outputs.}
Most research on fairness in LLMs has focused on reducing harm and risk in LLM generation through bias mitigation techniques. Techniques such as counterfactual data augmentation \citep{Lu2020GenderBias}, data filtering and selection \citep{Garimella2022DemographicAware}, machine unlearning methods \citep{yao-etal-2024-machine, dige2024mitigatingsocialbiaseslanguage}, designing specific prompting triggers \citep{Venkit2023} and incorporating the notion group fairness in constructing a bias evaluation dataset \citep{groupfairnesslens}, have proven effective to reduce stereotypical or harmful language targeted at various demographic groups. Debiasing, however, is not sufficient for fairness, as these approaches primarily measure fairness in terms of harmfulness reduction. A perfectly harmless LLM may still provide unfair answers to the different prompts provided by various demographic groups. 

\textbf{Aligning LLMs with Diverse Human Preferences.}
Recent work in fair RLHF, such as MaxMin RLHF \citep{MaxMinRLHF} and preference matching RLHF methods \citep{pmRLHF}, finetune the LLM to align with a diversity of human preferences. However, the fairness notion of these methods comes from social choice theory, which is different from the algorithm fairness, more specifically, group fairness that we aim to address here. In addition, these methods assume that people from various groups ask the same questions and, therefore, do not address the issue of diversity in informational needs.

\textbf{Group Fairness in LLM Decisions.}
Recent studies have explored the issue of group fairness when prompting LLMs to perform high-stakes ML classification decisions \citep{fairnesschatgpt, LLM_classifiers}. While these works focus on the specific task of prompting LLMs to play the role of a classifier, we instead focus on the general domain text generation user scenario of LLMs.


\section{Group Fairness in LLM}


We start to consider a particular definition of group fairness,  demographic parity (alternatively known as statistical parity) in the context of LLMs. First, we provide the group fairness of reward model definition for reward models. Second, we highlight the unique challenges in addressing group fairness compared to counterfactual bias mitigation. In addition, we outline the RLHF training pipeline and emphasize the importance of addressing group fairness in the reward model. 

\subsection{Group Fairness in Reward Models}

To define group fairness in reward models, we first present the definitions for social groups and protected groups.

\textbf{Definition 1 (Social Group) }
A social group \( G \subseteq \mathbb{G}\) is the population that shares an identity trait, which may be fixed, contextual, or socially constructed. Examples include demographic attributes collected through the census, including age, gender, and occupation.

\textbf{Definition 2 (Protected Attribute) }
A protected attribute is the shared identity trait that determines the group identity of a social group.


\textbf{Definition 3 (Group Fairness of Reward Models) }
Consider a model $\mathcal{M}$ that evaluates the quality of generated outputs from an LLM. Assume we have access to a set of prompts $X_G$, where the ground-truth quality of each prompt $x \sim X_G$ is equal. Let $\mathbb{E}_{x \sim X_G}[\mathcal{M}(x; \theta)]$ be the outcome measured by the reward model given a distribution of prompts $X_G$ specific to group $G \in \mathcal{G}$, where $\mathcal{G}$ represents a set of social groups, and each group $G$ has a different distribution of prompts $X_G$. Group fairness requires (approximate) parity in the average reward scores across all groups $G \in \mathcal{G}$, up to $\epsilon$, as measured by the reward model $\mathcal{M}$:
\begin{equation}
\left| \mathbb{E}_{x \sim X_G}[\mathcal{M}(x; \theta)] - \mathbb{E}_{x \sim X_{G'}}[\mathcal{M}(x; \theta)] \right| \leq \epsilon.
\end{equation}

\subsection{Challenges in Addressing Group Fairness with Bias Mitigation Techniques}

The evaluation and mitigation of counterfactual bias, often operationalized by switching group attributes (e.g., gender) at the prompt level, is a prevalent approach in assessing the fairness of large language models (LLMs). Fairness under these methods exists when the LLM's output for either prompt with switched attributes is the same. However, counterfactual bias evaluation in LLMs, as illustrated in Figure \ref{Conceptual Comparison}, inherently relies on assumptions that do not hold in real-world use scenarios.

\textbf{Uniformity of User Prompts Across Social Groups.} Current methods assume that users from different social groups will ask identical questions. When the prompts are inherently different questions, we can no longer substitute the protected attributes to measure fairness by verifying the outputs from LLM are the same.

\textbf{Explicit Inclusion of Group Attributes in Prompts.} This approach assumes that users will explicitly include their social group attributes (such as gender) in their prompts. In practice, users rarely identify their social group characteristics when writing prompts to interact with LLMs. 

These assumptions limit the method’s capacity to address rigorous concepts of algorithmic fairness. For instance, counterfactual bias evaluation does not fully adhere to the counterfactual fairness definition \citep{kusner2018counterfactualfairness}, as it omits the crucial concept of latent background variables. Therefore, it does not benefit from the equivalence between counterfactual fairness and group fairness as showed in traditional classification settings \citep{rosenblatt2023counterfactual}. Moreover, models that ignore protected attributes can achieve zero counterfactual bias by generating the same output, tending towards a definition of fairness through unawareness, which is a weaker definition than group fairness.

A potential alternative is outlined in Figure~\ref{Conceptual Comparison}. In this work, we contend that benchmarking the group fairness of reward models is a crucial first step toward developing LLMs that equitably serve all demographic groups, particularly given the reward model’s pivotal role in the RLHF pipeline.





\subsection{Sources of Bias in the RLHF Pipeline}

The RLHF pipeline typically involves three key stages: supervised fine-tuning, reward modeling, and reinforcement learning.

\textbf{Stage 1: Supervised Finetuning (SFT).}
In the first stage, a pre-trained language model is fine-tuned using supervised learning on task-specific datasets, such as dialogue, summarization, or instruction following, to create a reference policy denoted as $\pi_\text{ref}$.

\textbf{Stage 2: Reward Modeling.}
The second stage, reward modeling, seeks to capture human preferences of LLMs responses. Let $x$ be a prompt given to an LLM and $y$ be the model's output response for the prompt. For each given input $x$, LLM will generate a pair of responses and human annotators are asked to express their preference between two output responses, with $y_0$ and $y_1$ denote the chosen and rejected responses respectively. These human preference data are used to train a reward model $r_\theta(x, y)$, which learns to predict which response is better according to human judgment. Formally, the reward model’s loss derived from the Bradley-Terry (BT) preference model~\citep{bradley1952rank} can be expressed as:
\begin{equation}
\text{loss}(r_\theta) = -\mathbb{E}_{(x, y_0, y_1) \sim D} \left[ \log\left( \sigma \left( r_\theta(x, y_0) - r_\theta(x, y_{1}) \right) \right) \right],
\end{equation}
where $\sigma$ is the logistic function, and $D$ is the dataset of human-annotated preferences.

\textbf{Stage 3: Reinforcement Learning.}
Finally, in the third stage, the learned reward model is used in reinforcement learning to further optimize the model denoted as $\pi_{\phi}$, where $\phi$ is the weights of the LLM. The policy is trained to maximize the reward from the human feedback model while controlling for divergence from the initial supervised policy. The objective function of the reinforcement learning stage is usually given by:
\begin{equation}
    \max_{\phi} \mathbb{E}_{y \sim \pi_{\phi}(\cdot | x)} r(x, y) - \beta D_{\mathrm{KL}}(\pi_{\phi}(y | x) \| \pi_{\text{ref}}(y | x)),
\end{equation}
where $\beta$ controls the learned policy's deviation from the pretrained LLM as an initial reference policy $\pi_{\text{ref}}$. 

While all three stages can potentially introduce group unfairness into the final output of LLMs, this work focuses on the unfairness in the reward modeling stage. The reward models learned in this stage likely exhibit unfairness since neither the human preference dataset nor the Bradley-Terry model explicitly accounts for group fairness. Arguably, such unfairness in the reward model could be introduced to the final finetuned LLM after using it to train the LLM policy in the third stage.

\section{Benchmarking Reward Models}

\subsection{Constructing the Evaluation Dataset from The arXiv Metadata}

The arXiv Metadata dataset, which use is under the Creative Commons CC0 1.0 Universal (Public Domain Dedication) license, offers significant advantages to our fairness study. The dataset primarily consists of titles and abstracts from expert-written papers. The expert authorship ensures that the abstracts are high in quality, therefore receiving full scores on attributes such as correctness and coherence should be a minimum requirement. The reward model that satisfies group fairness should consistently deliver equal average reward scores for prompts and responses across all social groups.


\textbf{Selecting Social Groups.}
ArXiv papers are authored by experts across diverse fields.  Identifying social groups by occupation, such as physicists, economists, and computer scientists, we define eight demographic groups based on their disciplines: physics, mathematics, computer science, economics, electrical engineering, system science, quantitative biology, and quantitative finance.


\textbf{Evaluation Prompts and Responses.} 
We use expert-written texts from arXiv Metadata to benchmark group fairness in reward models. Each paper’s title and abstract form an evaluation pair: the prompt is generated as “Write an abstract for a paper with title <Title>”, and the expert abstract serves as the ground-truth response. A fair reward model should yield equal average scores across all eight categories.

Since the original arXiv Metadata dataset includes 200,000 papers, with fewer than 400 in the economics category, we use the arXiv API to collect more balanced data. We only include papers listed under a single category to avoid overlaps between groups, curating 2000 title-abstract pairs per category.



\subsection{Experimental Setup}

\textbf{Simplifying the Distributions of Prompts.}
To simplify the evaluation, we only do inference on prompts and responses that are unique to a specific group, assuming other groups never raise these questions as prompts to LLMs. In addition, we assume the distribution of prompts that all groups share is the same, therefore we are not evaluating on these shared common prompts as they will not affect the difference in group mean.

\textbf{Models.}
We only include reward models that can compute a reward score based on a single prompt and response message. LLM-as-a-Judge~\citep{zheng2024judging} and pairwise reward models are not included, as they require comparing two messages. The following 8 models from the RewardBench~\citep{rewardB} are selected in the evaluation: GRM-llama3-8B-sftreg~\citep{yang2024regularizing}, ArmoRM-Llama3-8B-v0.1~\citep{ArmoRM,wang2024arithmetic}, Eurus-RM-7b~\citep{yuan2024advancing}, FsfairX-LLaMA3-RM-v0.1~\citep{dong2023raft,xiong2024iterative}, Mistral-RM-for-RAFT-GSHF-v0~\citep{dong2023raft,xiong2023gibbs}, RM-Mistral-7B~\citep{dong2023raft,xiong2024iterative}, Nemotron-4-340B-Reward~\citep{wang2024helpsteer2}, and tulu-v2.5-13b-preference-mix-rm~\citep{ivison2024unpacking}.

\textbf{Recourses for Model Inference.} 
For the evaluation of the models, we utilized two NVIDIA A100 GPUs with 80 GB of memory for the tulu-v2.5-13b-preference-mix-rm model. API calls were employed for the Nemotron-4-340B-Reward model, leveraging external compute resources. For models with fewer than 8 billion parameters, such as GRM-llama3-8B-sftreg and ArmoRM-Llama3-8B-v0.1, we used NVIDIA RTX 6000 GPUs. Each model’s evaluation was completed within a maximum compute time of 3 hours.

\subsubsection{Group Fairness Metrics}

\textbf{Normalized Maximum Group Difference.}
The reward models are not trained to predict scores on the same scale. Therefore, directly computing the difference in group means is not a fair comparison. With this in mind, we propose a normalized maximum group difference score as a metric for group fairness. For each reward model, we compute the maximum difference in average rewards between any two social groups. This difference is then normalized by dividing it by the mean of the reward scores across all social groups.

\textbf{ANOVA as a Group Fairness Metric.}
To rigorously assess group fairness in the performance of reward models, we employ Analysis of Variance (ANOVA) as a statistical method to determine whether there are statistically significant differences between the means of rewards across different demographic groups defined in our study. ANOVA is instrumental in identifying whether variations in reward scores are due to inherent differences among the groups or are a result of random variations. This is critical in our context as it helps ensure that any observed difference in reward outcomes are attributable to the model’s unfairness across different groups.

\begin{table}[ht]
\centering
\small
\caption{ANOVA results for various reward models, assessing the significance of group differences in rewards.}
\vspace{0.8em}  
\begin{tabular}{lccc}
\toprule
\textbf{Reward Model} & \textbf{F-Statistics} & \textbf{p-Value} & \textbf{RewardBench Rank} \\
\midrule
ArmoRM-Llama3-8B-v0.1 & \textbf{70.44} & $9.46 \times 10^{-101}$ & 2 \\
GRM-llama3-8B-sftreg & 134.63 & $1.75 \times 10^{-193}$ & 8 \\
Eurus-RM-7b & 156.11 & $5.15 \times 10^{-224}$ & 16 \\
FsfairX-LLaMA3-RM-v0.1 & 232.98 & \(< 1 \times 10^{-300}\) & 12 \\
RM-Mistral-7B & 270.06 & \(< 1 \times 10^{-300}\) & 22 \\
tulu-v2.5-13b-preference-mix-rm & 384.86 & \(< 1 \times 10^{-300}\) & 19 \\
Nemotron-4-340B-Reward & 427.88 & \(< 1 \times 10^{-300}\) & 1 \\
Mistral-RM-for-RAFT-GSHF-v0 & 518.15 & \(< 1 \times 10^{-300}\) & 23 \\
\bottomrule
\end{tabular}
\vspace{0.8em}  
\label{anova_results}
\end{table}

\subsection{Results Analysis}

\begin{figure}[ht]
\centering  
\makebox[\linewidth]{%
    \includegraphics[width=1.1\linewidth]{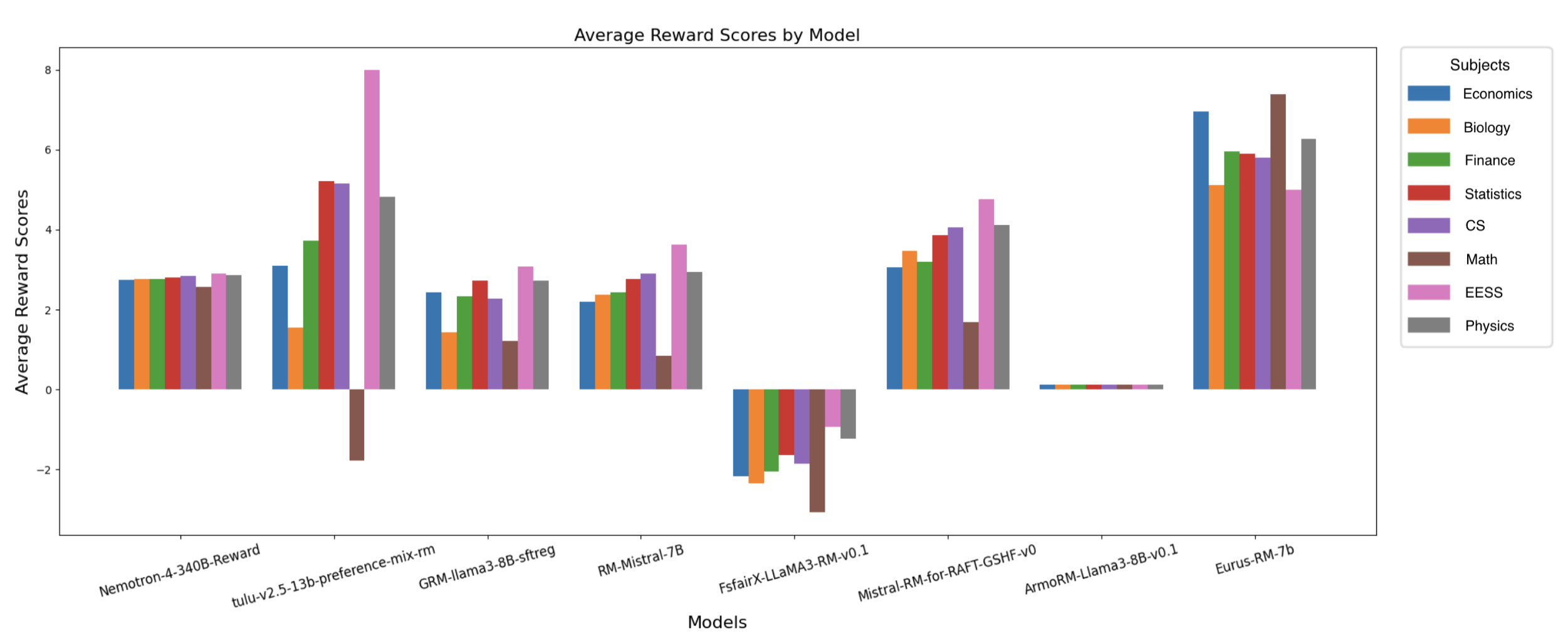}
}
\caption{Average Reward Scores by Model and Subject across various domains.}
\label{rewards by model}
\end{figure}


The plot for the average reward score of the selected 8 top-performing reward models from RewardBench is shown in Figure \ref{rewards by model}. Notice that not all reward models are on the same scale. For example, in the model design of ArmoRM-Llama3-8B-v0.1, a gating layer is applied to the outputs of the regression layer, resulting average rewards for all social groups close to zero.

Through a thorough analysis of the experiment results, we have made the following conclusions:

\textbf{The group unfairness in all reward models is statistically significant.}
Table \ref{anova_results} shows that every reward model has an F-statistic above 70 and a p-value below 0.0001, confirming substantial differences in group means. For example, ArmoRM-Llama3-8B-v0.1, the second highest ranked model on RewardBench, has the lowest F-statistic of 70.44, which is still well above 1 (the value indicating no group difference). Similarly, the Nemotron-4-340B-Reward model, despite its low normalized maximum group difference, has the second highest F-statistic, suggesting low within-group variance and significant group differences. These findings demonstrate that the disparities are systematic rather than random. 


\textbf{The best performing reward models are the fairer reward models.}
To compare the group fairness in the reward models, the normalized maximum group difference is computed. The results are shown in percentages in Table \ref{normalized maximum group difference}. The top 2 models from RewardBench Leaderboard, namely NemoTron-4-340B-reward and ArmoRM-Llama3-8B-v0.1 exhibit smaller Normalized Maximum Group Differences, substantially outperforming other models evaluated in this study, suggesting that the best reward models also exhibit the better group fairness.

\begin{table}[ht]
\centering
\small
\caption{Multiple Comparison of Means by the Tukey HSD Test}
\vspace{1em}  
\begin{tabular}{lc}
\toprule
\textbf{Reward Model} & \textbf{Significant Pairs / Total Pairs} \\
\midrule
GRM-llama3-8B-sftreg & \textbf{23 / 28} \\
ArmoRM-Llama3-8B-v0.1 & \textbf{23 / 28} \\
Eurus-RM-7b & 24 / 28 \\
FsfairX-LLaMA3-RM-v0.1 & 26 / 28 \\
Mistral-RM-for-RAFT-GSHF-v0 & 26 / 28 \\
RM-Mistral-7B & 25 / 28 \\
Nemotron-4-340B-Reward & 24 / 28 \\
tulu-v2.5-13b-preference-mix-rm & 25 / 28 \\
\bottomrule
\end{tabular}
\vspace{1em}  
\label{significant_pairs_table}
\end{table}

\begin{table}[ht]
\centering
\small
\caption{Differences in average rewards between the maximum and minimum values for each reward model, expressed as percentages. The score with the lowest absolute value is in bold.}
\vspace{0.8em}  
\begin{tabular}{lcc}
\toprule
\textbf{Model} & \textbf{Normalized Maximum Group Difference} & \textbf{RewardBench Rank} \\
\midrule
Nemotron-4-340B-Reward & 12.49\% & 1 \\
tulu-v2.5-13b-preference-mix-rm & 262.89\% & 19 \\
GRM-llama3-8B-sftreg & 82.09\% & 8 \\
RM-Mistral-7B & 110.63\% & 22 \\
FsfairX-LLaMA3-RM-v0.1 & -111.52\% & 12 \\
Mistral-RM-for-RAFT-GSHF-v0 & 87.46\% & 23 \\
ArmoRM-Llama3-8B-v0.1 & \textbf{9.78\%} & 2 \\
Eurus-RM-7b & 39.53\% & 16 \\
\bottomrule
\end{tabular}
\vspace{0.8em}  
\label{normalized maximum group difference}
\end{table}

\begin{figure}[ht]
\centering  
\makebox[\linewidth]{%
    \includegraphics[width=1.1\linewidth]{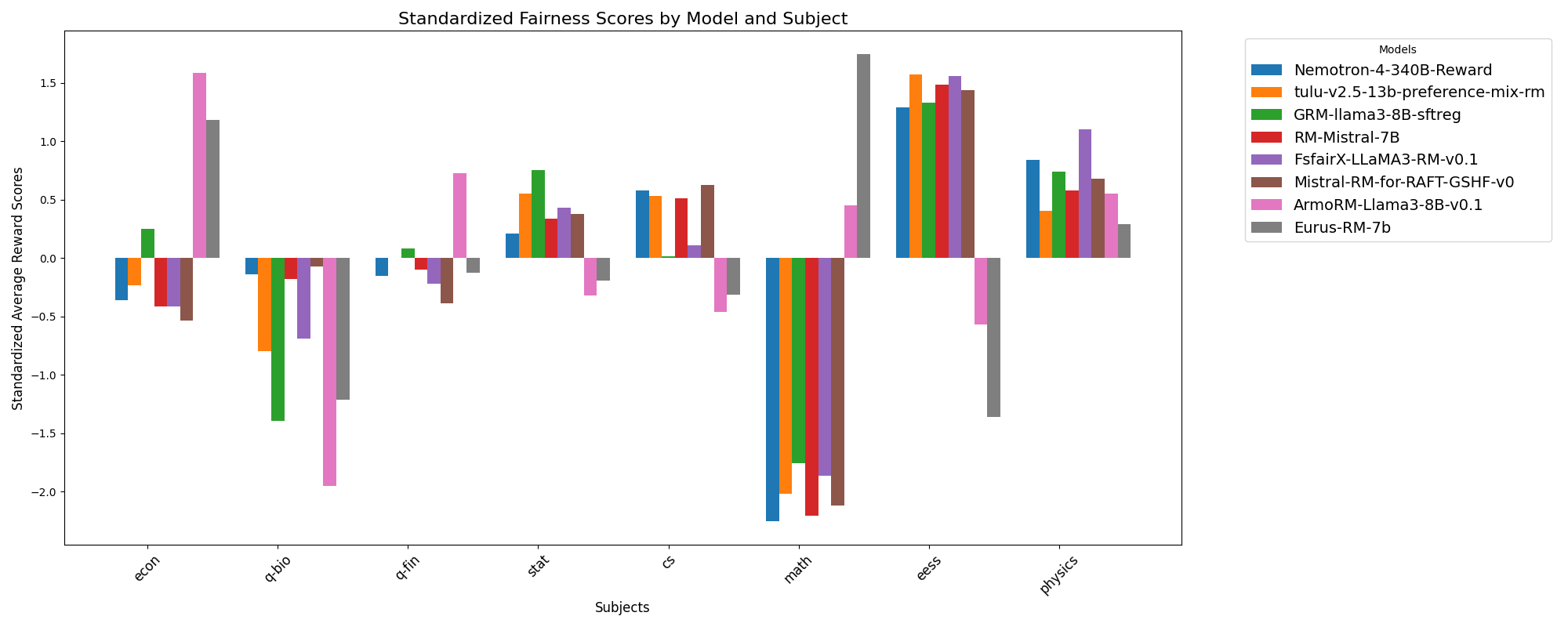}
}
\caption{Fairness Scores by Model and Subject across various domains.}
\label{rewards by subject}
\end{figure}


\textbf{Group unfairness exists in most pairs of demographic groups in every reward model.} The Tukey HSD Test, a post-hoc Analysis of ANOVA in Table \ref{significant_pairs_table}, shows that each reward model has at least or more than 23 pairs of groups that shows significant differences in the group mean out of a total of all 28 possible combinations of pairs for 8 groups. This indicates that the significant findings from ANOVA are not a result of a significant difference between a only few groups, but rather widespread differences in group means across the majority of group comparisons.

\begin{table}[ht]
\centering
\small
\caption{Pearson Correlation Coefficients of NVIDIA Nemotron Model with Other Models}
\label{tab:pearsonr}
\vspace{0.8em}  
\begin{tabular}{lc}
\toprule
\textbf{Model} & \textbf{Pearson Correlation Coefficient} \\
\midrule
tulu-v2.5-13b-preference-mix-rm & 0.942 \\
RM-Mistral-7B & 0.991 \\
Mistral-RM-for-RAFT-GSHF-v0 & 0.988 \\
FsfairX-LLaMA3-RM-v0.1 & 0.945 \\
GRM-llama3-8B-sftreg & 0.820 \\
Eurus-RM-7b & -0.738 \\
ArmoRM-Llama3-8B-v0.1 & -0.255 \\
\bottomrule
\end{tabular}
\vspace{0.8em}  
\end{table}


\textbf{A systematic unfairness might exist in reward models.}
To elucidate the variations in average rewards across different demographic groups, we present a standardized comparison of average rewards by subject in Figure \ref{rewards by subject}. This analysis reveals a consistent pattern of disparate treatment for all demographic groups across most reward models. For a better illustration, besides ArmoRM-Llama3-8B-v0.1 and Eurus-RM-7b, the 340B Nemotron model exhibits a Pearson correlation of over 0.8 with all of the rest reward models (in some cases 0.99), as shown in Table \ref{tab:pearsonr}. The congruence in average reward score disparities across the majority of models suggests a systemic bias that may originate from similar methodologies in their training datasets and algorithms.

\section{Conclusion}

In this work, we introduced a new problem of group fairness in reward models as the first step to address the challenge of creating large language models (LLMs) that benefit all groups of users equitably. Our proposed benchmark reveals significant and pervasive unfairness across various reward models, highlighting the need for unfairness mitigation in reward models. We conduct extensive quantitative experiments on eight top-performing reward models, using a novel dataset derived from arXiv metadata. The results demonstrate the effectiveness of our approach in identifying group unfairness and suggest a correlation between model performance and fairness. This work lays the foundation for developing more equitable AI systems and opens new directions for group fairness research in LLMs.

\subsubsection*{Acknowledgments}
This work was supported in part by the US National Science Foundation under grants III-2128019 and SLES-2331904. 
This work was also supported in part by the Coastal Virginia Center for Cyber Innovation (COVA CCI) and the Commonwealth Cyber Initiative (CCI), 
an investment in the advancement of cyber research $\text{\&}$ development, innovation, and workforce development. For more information about CCI, visit 
\url{www.covacci.org} and \url{www.cyberinitiative.org}.


\bibliography{main}
\bibliographystyle{rlj}


\newpage
\appendix


\newpage





\end{document}